\documentclass{article}

\usepackage{microtype}
\usepackage{graphicx}
\usepackage{subcaption}
\usepackage{booktabs} 

\usepackage{hyperref}

\usepackage[accepted]{icml2026}

\usepackage{amsmath}
\usepackage{amssymb}
\usepackage{mathtools}
\usepackage{amsthm}

\usepackage{listings}
\usepackage{xcolor}

\lstdefinestyle{promptstyle}{
  basicstyle=\ttfamily\scriptsize,
  breaklines=true,
  breakatwhitespace=false,
  columns=fullflexible,
  keepspaces=true,
  frame=none,
  showstringspaces=false,
  postbreak=\mbox{\textcolor{gray}{$\hookrightarrow$}\space},
}

\usepackage[capitalize,noabbrev]{cleveref}

\theoremstyle{plain}

\theoremstyle{definition}

\theoremstyle{remark}

\usepackage[textsize=tiny]{todonotes}

\icmltitlerunning{Solving Is Not Drawing: A Benchmark for Diagrammatic Reasoning in Olympiad Geometry}

\begin{document}

\twocolumn[
  \icmltitle{Solving Is Not Drawing: A Benchmark for Diagrammatic Reasoning in Olympiad Geometry}



  \icmlsetsymbol{equal}{*}

\begin{icmlauthorlist}
    \icmlauthor{Hsien Xin Peng}{equal,algo}
    \icmlauthor{Anthony Kim}{equal,algo}
    \icmlauthor{Alvin Li}{equal,algo}
    \icmlauthor{Calvin Supasanya}{equal,algo}
    \icmlauthor{Shivank Garg}{algo}
    \icmlauthor{Kevin Zhu}{algo}
  \end{icmlauthorlist}

  \icmlaffiliation{algo}{Algoverse AI Research}
  \icmlcorrespondingauthor{Shivank Garg}{shivank@algoverseairesearch.org}

  \icmlkeywords{AI for math, ICML}
  \vskip 0.3in
]



\printAffiliationsAndNotice{}  

\begin{abstract}
Foundation models such as GPT and Claude now solve olympiad-level mathematics with remarkable proficiency, so much so that geometry problem solving has become a standard proxy for their mathematical reasoning. Yet solving a geometry problem and drawing the figure it depends on are not the same skill: progress often hinges on a faithful diagram with the right auxiliary constructions and incidences, and it is unclear that a model which reasons its way to the answer can also produce one. A growing collection of benchmarks, including MathVista, and MathVerse, measures whether models reach the correct answer, but to our knowledge, none isolate the distinct ability to construct the diagram itself, leaving this capability unmeasured. We introduce an open-source benchmark that targets this gap: 954 self-contained olympiad geometry problems, with a 297-problem hard subset, each paired with its solution and a human-authored, high-fidelity diagram in renderable Asymptote code, together with a suite of text-, code-, image-, VLM-, and constraint-based metrics for what we term diagrammatic reasoning. Evaluating current foundation models reveals a pronounced gap between solving and drawing: their diagrams are markedly less faithful, with an average compile success rate of only 36.14\%. Strong mathematical reasoning, we find, does not imply the ability to construct accurate geometric diagrams. Our benchmark and dataset can be accessed at \url{https://huggingface.co/datasets/max98765/hard_geometry_problems_with_diagrams}. 
\end{abstract}

\section{Introduction}
Recently, foundation models have achieved exceptional scores on math
olympiad problems, so much so that these problems are now used as a
general proxy for a model's mathematical reasoning. However, many
olympiad problems rely on more than textual reasoning. Geometry is a
prominent example, as progress frequently depends on an accurate,
representative diagram, complete with the right auxiliary constructions
and the correct incidences between points, lines, and circles. Solving
the problem and drawing the figure it rests on are not obviously the
same skill.

Our experiments show that they are not. Despite their textual
problem-solving ability, foundation models exhibit a clear discrepancy
between solve accuracy and faithful diagram generation, and this gap
persists even when a model is handed the full reference solution and
asked only to render a consistent figure. Since geometric construction
and visualization are central to solving geometry problems, we define
\textit{diagrammatic reasoning} as a model's ability to produce a
geometrically consistent, high-fidelity diagram from a problem, and we
treat it as a capability distinct from problem solving. Measuring it
matters in practice: a model that reasons correctly but draws an
inconsistent figure cannot be trusted to support geometric work, to
verify its own constructions, or to feed reliable diagrams to
downstream agents.

A growing collection of benchmarks evaluates the mathematical and
geometric problem-solving abilities of foundation models, including
OlympiadBench, MathVista, MathVerse, and We-Math. These efforts measure
whether a model reaches the correct answer or follows a valid reasoning
trace, but they consume diagrams as input or treat them as incidental.
To our knowledge, none isolate the distinct ability to construct the
diagram itself, so a model can score well while holding an inconsistent
or impossible picture of the configuration. 

We address this gap with an open-source benchmark of 954 self-contained
olympiad geometry problems, including a 297-problem hard subset, each
paired with its full solution and a human-authored, high-fidelity
ground-truth diagram available as both an image and renderable code. We
require models to express diagrams as Asymptote code, since its native
geometric primitives make it better suited to precise compass-and-
straightedge constructions than general-purpose plotting libraries,
while keeping the task in the model's native text modality and yielding
figures that are precise, modifiable, and easy to compare against
ground truth diagrams. That the gap persists in this code setting
indicates it is not an artifact of weak multimodal generation; even
when confined to text, models struggle to extract and correctly represent
geometric constraints. We pair the dataset with a suite of text-,
code-, image-, VLM-, and constraint-based metrics for diagrammatic
reasoning.

Evaluating current foundation models on this benchmark confirms a
pronounced gap between solving and drawing. Their diagrams are markedly
less faithful than their solve accuracy would suggest, with an average
compile success rate of only 36.14\% and fidelity well below their
problem-solving performance, as reported in Table~\ref{tab:results}. To
summarize, our main contributions are as follows:
\begin{enumerate}
    \item We define and standardize \textit{diagrammatic reasoning},
    the ability to generate a geometrically consistent, high-fidelity
    diagram, as a capability distinct from problem solving.
    \item We release an open-source dataset of 954 olympiad geometry
    problems, each with a full solution and a human-made, high-fidelity
    diagram available as both an image and renderable Asymptote code,
    together with a suite of text-, code-, image-, VLM-, and
    constraint-based metrics for evaluation.
    \item We benchmark state-of-the-art foundation models and show
    that strong mathematical reasoning does not imply the ability to
    construct accurate geometric diagrams.
\end{enumerate}

\section{Related Works}

\subsection{Geometric Diagram Generation}
A range of approaches exist for generating accurate geometry diagrams.
Manual tools such as GeoGebra \cite{hohenwarter2007geogebra} let users
construct figures directly, while code-based generation relies on
libraries such as matplotlib \cite{hunter2007matplotlib} or specialized
languages such as Asymptote \cite{asymptoteManual}; a separate line of
work defines formal geometry languages and uses solvers to satisfy
geometric constraints \cite{lu2021intergps,wang2025magicgeo,zhang2025geosdf}.
More recently, the field has shifted toward learning-based generation
\cite{wang2025magicgeo,zhang2025geosdf,cheng2025geouni}. Diffusion models
produce high-fidelity images \cite{ho2020ddpm,rombach2022latentdiffusion}
but tend to sacrifice the mathematical precision that geometry demands,
which has motivated Multi-modal Large Language Model (MLLM) approaches
that combine textual and visual modalities \cite{lu2024mathvista,cheng2025geouni}.
GeoUni \cite{cheng2025geouni}, for instance, trains a unified model with
GRPO \cite{shao2024deepseekmath} and several geometry-specific rewards
to generate accurate diagrams, but this requires training a dedicated
MLLM where a text- and code-based approach could suffice.

Foundation models such as GPT-5.4 \cite{openai2026gpt54} and Claude
Sonnet 4.6 \cite{anthropic2026sonnet46} are already strong textual and
mathematical reasoners, and can generate matplotlib or Asymptote code
that renders into a diagram without any dedicated training or additional
modality. Whether the diagrams rendered from this code are
geometrically faithful to the original problem has not, to our
knowledge, been systematically evaluated. This is the gap our work
targets. We determine whether an existing foundation model can produce
geometrically consistent, code-based diagrams.

\subsection{Geometry Metrics and Evaluations}
Prior geometry benchmarks typically measure whether a model reaches the
correct answer or produces a valid reasoning trace
\cite{chen2022geoqa,lu2021intergps,chen2022unigeo,lu2024mathvista}, but not whether a model conceptualizes the geometric structure
correctly; a model can reach the right answer through textual reasoning while holding an inconsistent diagrammatic representation \cite{zhang2024mathverse,qiao2025wemath}. Comparing a generated diagram against a ground-truth construction instead requires fidelity metrics.

For images, learned perceptual distances such as LPIPS \cite{zhang2018unreasonable} captures visual similarity. For code, text-overlap metrics such as BLEU \cite{papineni2002bleu} and chrF++ \cite{popovic2017chrf} count shared word and character n-grams and CodeBLEU \cite{ren2020codebleu} adds syntactic and semantic signals, though all compare token distributions rather than geometric behavior. Using an LLM or Vision-Language Model as a judge \cite{zheng2023judging,chen2024mllm} assesses geometric reasoning more directly, which we adopt alongside the metrics above.

\subsection{Geometry Datasets}
Existing geometry datasets focus on problem solving rather than diagram fidelity. For instance, GeoQA \cite{chen2022geoqa} and Geometry3K \cite{lu2021intergps} pair problems with diagrams and formal annotations for answer prediction; UniGeo \cite{chen2022unigeo} unifies calculation and proof, and MathVista \cite{lu2024mathvista}, MathVerse \cite{zhang2024mathverse}, and We-Math \cite{qiao2025wemath} target general visual mathematical reasoning. Separately, scientific vector graphics benchmarks show that code-based diagrams can serve as evaluation targets. For example, AutomaTikZ \cite{belouadi2024automatikz}, DeTikZify \cite{belouadi2024detikzify}, Text2arch \cite{garg2026textarch} and vTikZ \cite{reux2025vtikz} all focus on general diagrams rather than theorem-driven geometry, which we address. 


\section{Dataset Curation}

\begin{figure*}[htbp]
  \centering
  \includegraphics[width=\textwidth]{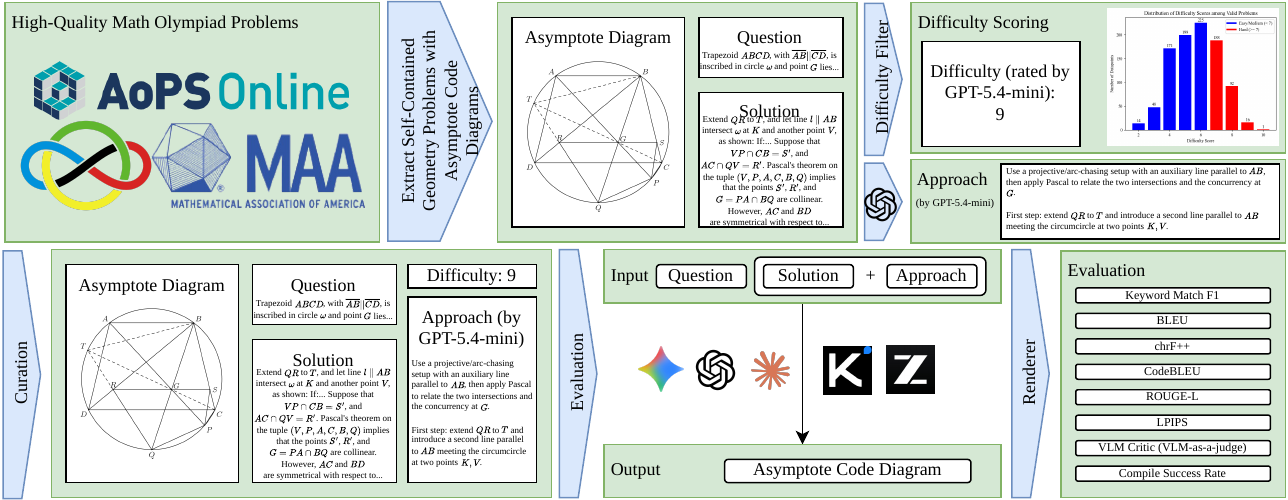}
  \caption{Overview of the Data Curation and Evaluation Processes. We web-scrape high-quality mathematical olympiad problems, namely from national and international competitions with well-documented solutions. We then extract human-made Asymptote code to generate the diagrams corresponding to each solution. To create our dataset, we filter our data, eliminating problems that are not self-contained or not geometric. We then prompt GPT-5.4-mini to generate a difficulty score and a summary of each solution's approach for further analysis. Using a difficulty cutoff, we curate a subset of harder geometry problems. We evaluate foundation models on a range of metrics with two different experimental settings: 1) a baseline with solely the problem statement and 2) including the full reference solution and approach.}
  \label{fig:diagram}
\end{figure*}

\paragraph{Problem Sources} We source problems from a range of mathematical olympiad exams, primarily the American Mathematics Competition (AMC), American Invitational Mathematics Exam (AIME), USA Math Olympiad (USAMO) and Junior Math Olympiad (USAJMO) \cite{maaAMC,maaInvitational}, and the International Mathematical Olympiad (IMO) \cite{imoOfficialProblems}. We web-scraped problems dating back to the 1980s and their online solutions \cite{aopsProblems}. We keep only solutions whose diagrams were generated with human-authored Asymptote code, which we treat as ground truth. We then prompted GPT-5.4-mini \cite{openai2026gpt54mini} to, for each problem, (1) determine whether it is geometric, (2) determine whether it was self-contained, and (3) summarize the reference solution's approach. In total, we curated a dataset of 954 problems, each paired with a human-authored, Asymptote code-based diagram, and isolate a subset of 297 problems as part of a more challenging dataset. An overview of the curation and evaluation pipeline is shown in Figure 1, and a full dataset breakdown appears in Appendix \ref{app:dataset-breakdown}.

\paragraph{Self-Containment} We define a problem as ``self-contained'' if its solution is fully reproducible from the problem statement alone, and we exclude any problem that depends on an input image. This serves two purposes. First, it keeps the benchmark accessible to text-only models. Second, it removes a source of error propagation, since a misread input diagram would corrupt a model's reasoning from the outset. A purely text-based input avoids both problems.


\paragraph{Difficulty Filtering} We prompted GPT-5.4-mini to rate each
problem's difficulty on a scale from 1 to 10. We then retained only problems scoring 7 or above, yielding a hard subset of 297 problems. We chose this cutoff for two reasons: it keeps the benchmark sufficiently challenging to be discriminative, and keeps it comparable in size to other evaluation benchmarks. A more detailed breakdown of the data is in Appendix \ref{app:dataset-breakdown}.

\section{Methodology}

\subsection{Models}
We evaluate Claude Sonnet 4.6, Gemini 3.1 Flash-lite, GPT-5.4-mini, Kimi K2.5, and GLM 5.1 on their ability to generate high-quality, code-based diagrams.

\subsection{Experiments}

We run two settings per model, differing in how much guidance the model receives toward our ground-truth solution. We initially considered providing only the problem statement, but since the problems admit many valid solution paths, a model may diverge from our ground truth and have its diagram unfairly penalized against a single reference; we therefore add a second setting in which the model is anchored to the intended construction via the full reference solution. In the with-solution setting, we provide the statement together with the full reference solution and prompt the model to render a consistent diagram, directly measuring whether it can produce a diagram faithful to a given solution. In the without-solution setting, we provide only the problem statement and prompt the model to both reason through the solution and draw it, jointly evaluating textual and geometric reasoning. Prompts are given in Appendix~\ref{app:prompts}.

\subsection{Metrics}
We evaluate models with a mix of text-, code-, image-, and VLM-based metrics that together capture code similarity and final diagram fidelity. Full details are given in Appendix~\ref{app:metrics}.

\paragraph{Text- and Code-Based Metrics}
We report a Keyword Match F1 score that cross-references key Asymptote functions in the generated code against those in the ground truth, along with a family of overlap metrics: BLEU~\cite{papineni2002bleu} and chrF++ \cite{popovic2017chrf} measure shared word and character n-grams, CodeBLEU \cite{ren2020codebleu} adds syntactic and semantic program signals, and ROUGE-L \cite{lin-2004-rouge} measures longest-common-subsequence overlap. These metrics are cheap and give a rough indication of how close a generated program is to the ground truth. They cannot, however, capture geometric relations between elements: a high score may reflect only shared boilerplate, while a correct diagram built from different Asymptote functions may be penalized. Since Asymptote code is highly formulaic, we retain them as coarse indicators of consistency rather than measures of geometric correctness.

\paragraph{Image-Based Metrics}
For visual fidelity, we use Learned Perceptual Image Patch Similarity (LPIPS)~\cite{zhang2018unreasonable}, which compares deep feature representations of the rendered diagrams; lower values indicate greater similarity.

\paragraph{VLM-Based Metrics}
We use GPT-5.4-mini as a vision-language judge, scoring each generated diagram against the ground truth on geometric similarity. The prompt is given in Appendix~\ref{app:prompts}.

\paragraph{Compilation Metrics}
We report a Compile Success Rate, the fraction of generated Asymptote programs that render without error, as a basic check of whether a model's output is usable at all.

\section{Results}
We present a comprehensive list of our results in Table \ref{tab:results}.

\begin{table*}[h!]
    \centering
    \scriptsize
    \setlength{\tabcolsep}{4pt} 
    \caption{Experimental Results (Mean Score Scaled to 100). LPIPS is lower-is-better. VLM Critic raw scores were on a 0--10 scale and have been linearly rescaled to 0--100 for consistency. ``With'' denotes that the model was given the full reference solution and approach as context; ``Without'' denotes that the model was given only the problem statement.}
    \label{tab:results}
    \begin{tabular}{lcccccccccc}
        \toprule
        \textbf{Method} & \multicolumn{2}{c}{\textbf{GPT5.4-mini}} & \multicolumn{2}{c}{\textbf{Claude Sonnet 4.6}} & \multicolumn{2}{c}{\textbf{Gemini 3.1 Flash-lite}} & \multicolumn{2}{c}{\textbf{Kimi K 2.5}} & \multicolumn{2}{c}{\textbf{Glm 5.1}} \\
        \cmidrule(lr){2-3}
        \cmidrule(lr){4-5}
        \cmidrule(lr){6-7}
        \cmidrule(lr){8-9}
        \cmidrule(lr){10-11}
        
        & \textbf{With} & \textbf{Without}
        & \textbf{With} & \textbf{Without}
        & \textbf{With} & \textbf{Without}
        & \textbf{With} & \textbf{Without}
        & \textbf{With} & \textbf{Without} \\
        \midrule
        \multicolumn{11}{l}{\textbf{Text/Code-Based Metrics}} \\
        \midrule
        
        $\uparrow$ Keyword Match F1    & 55.4 & 51.4 & 55.9 & 52.8 & 47.4 & 42.1 & 49.4 & 47.9 & 54.6 & 51.5 \\
        $\uparrow$ BLEU                & 18.5 & 16.8 & 15.7 & 15.3 & 16.6 & 13.9 & 19.1 & 16.4 & 17.8 & 16.3 \\
        $\uparrow$ chrF++              & 35.7 & 32.9 & 34.7 & 32.4 & 30.5 & 27.2 & 33.8 & 30.1 & 35.1 & 31.9 \\
        $\uparrow$ CodeBLEU       & 17.1 & 15.1 & 15.1 & 14.3 & 14.8 & 12.4 & 17.7 & 15.4 & 16.4 & 15.0 \\
        $\uparrow$ ROUGE-L             & 22.4 & 21.6 & 20.2 & 20.2 & 24.8 & 23.4 & 24.9 & 24.0 & 22.6 & 21.9 \\
        
        \midrule
        \multicolumn{11}{l}{\textbf{Image-Based Metrics}} \\
        \midrule
        
        $\downarrow$ LPIPS             & 54.0 & 53.7 & 50.6 & 51.3 & 52.6 & 52.7 & 47.9 & 49.5 & 50.8 & 53.3 \\
        \midrule
        \multicolumn{11}{l}{\textbf{VLM-as-a-Judge}} \\
        \midrule
        
        $\uparrow$ VLM Critic          & 37.9 & 46.4 & 49.5 & 51.0 & 42.6 & 45.2 & 50.8 & 53.2 & 44.2 & 45.0 \\
        \midrule
        \multicolumn{11}{l}{\textbf{Compilation Metrics}} \\
        \midrule
        
        $\uparrow$ Compile Success Rate & 21.7 & 20.8 & 27.6 & 41.3 & 59.2 & 59.3 & 30.9 & 22.0 & 40.0 & 38.6 \\
        \bottomrule
    \end{tabular}
\end{table*}

\begin{figure*}[t]
    \centering
    \begin{minipage}[t]{0.46\textwidth}
        \centering
        \includegraphics[width=\linewidth]{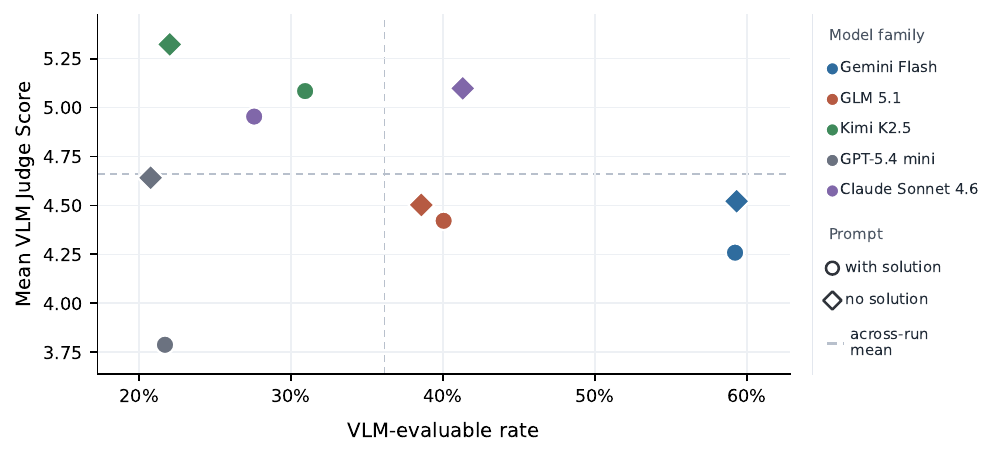}
        \caption{Mean VLM judge score against VLM-evaluable rate for each model and prompt setting (circle: with solution; diamond: no solution). Dashed lines mark the across-run means. Higher evaluable rates do not correspond to higher judge scores: the most reliably renderable models score below the mean on geometric fidelity, while the best-judged runs render less often.}
        \label{fig:vlm_scatter}
    \end{minipage}
    \hfill
    \begin{minipage}[t]{0.50\textwidth}
        \centering
        \includegraphics[width=\linewidth]{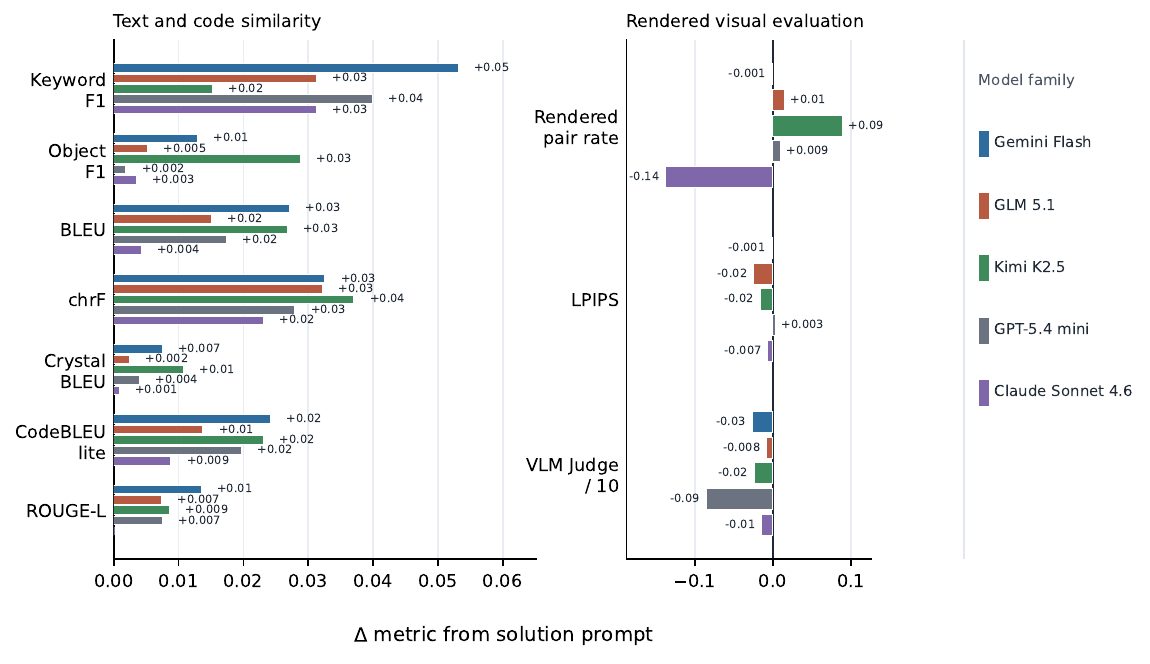}
        \caption{Change in each metric when the full reference solution is provided, relative to the no-solution setting (with minus without), per model. Left: text- and code-similarity metrics improve uniformly. Right: rendered-image metrics change little and often negatively, indicating that solution context raises code overlap without improving diagram fidelity.}
        \label{fig:delta_solution}
    \end{minipage}
\end{figure*}

\subsection{Solving Performance Does Not Predict Diagram Fidelity}
Across every model, diagram-fidelity scores fall well short of what the same models' problem-solving accuracy would suggest. The text- and code-based metrics remain low (BLEU between 13.9 and 19.1, CodeBLEU between 12.4 and 17.7), and the VLM critic, our most direct measure of geometric faithfulness, never exceeds 53.2 on a 0-100 scale. A model that solves a problem correctly can still produce a diagram sharing little geometric structure with the reference, confirming that diagrammatic reasoning is a capability distinct from textual problem solving rather than a by-product of it.

\subsection{Rendering More Is Not Rendering Better}
Compile success and geometric fidelity are only weakly aligned. Gemini 3.1 Flash-lite compiles most reliably (around 59\% in both settings) yet scores below the across-run mean on the VLM critic, whereas Kimi K2.5 and Claude Sonnet 4.6 earn the highest judge scores while compiling far less often. Figure~\ref{fig:vlm_scatter} illustrates that the models with the highest compile rates are not those with the highest judge scores. Compile rate alone is therefore a poor proxy for diagrammatic quality, since a model can produce renderable code while still drawing the wrong figure.

\subsection{Effect of Providing the Reference Solution}
Supplying the full reference solution improves surface-level code similarity but does little for rendered fidelity. Figure~\ref{fig:delta_solution} shows the per-metric difference between the with-solution and no-solution settings. Every text- and code-similarity metric improves modestly and consistently with the solution, indicating that models reuse recognizable structure from the provided text. The rendered-image metrics, by contrast, change little and often negatively. The solution context shifts the generated code toward the reference text without making the resulting diagram more geometrically correct, which places the bottleneck in construction rather than in understanding the solution.

\subsection{Comparison with Baseline Performance}
Due to compute constraints, we did not re-run the models on the problem-solving task. Instead, we report each model's published accuracy on AIME 2025, a competition that contributes a large share of our dataset, as drawn from the respective system cards and reputable secondary sources. On this benchmark the evaluated models are near-saturated: the GPT-5 line reports up to $94.6\%$ without tools and a perfect $100\%$ with code execution \cite{openai2025gpt5}, and the Claude Sonnet line similarly reports roughly $87\%$ without tools and $100\%$ with tools \cite{anthropic2026sonnet46}, with comparable figures across the other models we evaluate. Placing these solving accuracies beside the diagram-fidelity results in Table~\ref{tab:results} exposes a large and
consistent gap: models that solve the underlying problems almost perfectly produce diagrams whose VLM-critic fidelity remains below $54/100$ and whose programs compile only $36.14\%$ of the time on average. Foundation models thus reliably solve textual geometry while falling short of constructing the accurate, high-fidelity diagrams that geometric reasoning depends on.

\section{Conclusion}
In this paper, we address the issue of diagrammatic reasoning in existing foundation models. While these models are exceptional at mathematical reasoning tasks, we demonstrate that they fail to generate diagrams that are consistent with textual solutions, even when those solutions are provided to them. These skills are highly pertinent to problems that require geometric reasoning and awareness. To measure it, we introduce a high-quality benchmark of olympiad geometry problems, each paired with a human-authored solution and a ground-truth diagram in renderable Asymptote code, together with a suite of text-, code-, image-, and VLM-based metrics. Benchmarking current foundation models on this dataset provides, to our knowledge, the first concrete evaluation of their diagrammatic reasoning, and reveals a pronounced gap between solving and drawing: generated diagrams achieve an average compile success rate of just 36.14\% and an average VLM-critic score of 46.58\%, far below the models' problem-solving ability. These results highlight a continuing need to bring foundation models' diagrammatic reasoning and geometric awareness up to par with their textual reasoning.

\section{Limitations}
Due to compute constraints and high API costs associated with model inference we did not run the evaluated models on the problem-solving task ourselves, and instead drew baseline accuracies (such as AIME 2025 performance) from published system cards. For the same reason, we generated only a single diagram per problem from each model rather than sampling multiple times; because generation is stochastic, our reported compile-success and fidelity scores reflect one sample per problem and may vary across runs. Our automatic metrics are not perfect proxies for geometric correctness: text- and code-overlap scores reward shared Asymptote boilerplate, and metrics such as SymPy-based symbolic match, could improve evaluation reliability. Our most direct fidelity measure relies on a VLM judge, which inherits the judge model's own visual and geometric limitations and may misjudge borderline diagrams.





\bibliography{example_paper}
\bibliographystyle{icml2026}

\newpage
\appendix
\onecolumn

\section{Prompts}
\label{app:prompts}

\subsection{Experiments}
For our evaluations of foundation models, we use with the following system prompt:

\begin{lstlisting}[style=promptstyle]
You are an expert data visualization programmer specializing in research-quality diagrams using Asymptote.
Generate a complete, self-contained Asymptote script that creates a diagram based on the user's request.
Rules:
1. Use Asymptote (asy) for all diagrams. Import packages as needed, e.g.:
   - import graph;   // for 2-D function/data plots
   - import stats;   // for histograms, bar charts
   - import three;   // for 3-D diagrams
2. Do NOT specify an output file inside the script. The output path is passed via the -o CLI flag.
3. Set canvas size with size(width, height) or size(300) at the top of the script.
4. Follow research publication standards:
   - Clear axis labels with units where applicable
   - Descriptive title using label() or xaxis()/yaxis() labels
   - Legible font sizes
   - Professional colors (use Asymptote named colors or rgb() values; avoid garish colors)
   - Include a legend when multiple series are present
5. Use synthetic/example data if the user does not provide specific data. Make it realistic and meaningful.
6. IMPORTANT: cos() and sin() take RADIANS. Use dir(degrees) for unit vectors at a given angle in degrees.
7. IMPORTANT: label() only accepts ONE pen argument. Combine pens with +: label("$x$", p, blue+fontsize(10pt)) not label("$x$", p, fontsize(10pt), blue).
8. Output ONLY the Asymptote code inside a single ```asy code block. No other text.
\end{lstlisting}

We then follow up with the following user prompt for evaluations in the with-solution setting.
\begin{lstlisting}[style=promptstyle]
Create a diagram for:

\{question\}

Solution:

\{solution\}

**Approach to use:**

\{approach\}

\end{lstlisting}

For evaluations without a solution and approach, we use this user prompt instead:
\begin{lstlisting}[style=promptstyle]
Create a diagram for:

\{question\}

\end{lstlisting}

\subsection{VLM Critic}
For our VLM-as-a-judge metric, we use GPT-5.4-mini as our judge. We used the following prompt and appended the two rendered images (the ground truth image and the generated image) to create our full prompt.

\begin{lstlisting}[style=promptstyle]
You are evaluating geometry diagram generation for a research benchmark.
You will see two rendered images:
1. The reference image rendered from the ground-truth Asymptote code.
2. The generated image rendered from a model's Asymptote code.

Judge geometric similarity, not artistic style. Focus on whether the generated
diagram contains the same geometric objects, labels, incidences, relative
layout, angle/parallel/perpendicular markings, and important construction
elements as the reference. Ignore minor rendering artifacts such as antialiasing,
line thickness, small font differences, and harmless scaling/cropping.

Return only JSON with these fields:
{
  "score_0_10": number,
  "summary": string,
  "missing_elements": list[string],
  "wrong_geometry": list[string],
  "extra_elements": list[string],
  "label_issues": list[string]
}

Scoring:
- 9-10: essentially the same geometry and labels.
- 7-8: mostly correct with minor missing or misplaced details.
- 5-6: recognizable core construction but several meaningful issues.
- 3-4: major geometry mismatch, but some relevant structure remains.
- 1-2: mostly wrong or unreadable.
- 0: blank, non-diagram, or completely unrelated.
\end{lstlisting}

\section{Metrics}
\label{app:metrics}
\paragraph{Text- and Code-Based Matching Metrics}
We use a Keyword Match F1 score that cross-references key Asymptote functions present in the generated code against those in the ground truth. Alongside it we report a family of overlap metrics: BLEU, chrF++, CodeBLEU, and ROUGE-L. At its core, BLEU measures n-gram precision according to:

\begin{equation}
\mathrm{BLEU}
=
\mathrm{BP}
\cdot
\exp\left(
\sum_{n=1}^{N}
w_n \log p_n
\right)
\end{equation}

where $w_n$ are the weights and $\mathrm{BP}$ is the brevity penalty,
\begin{equation}
\mathrm{BP}
=
\begin{cases}
1 & \text{if } c > r \\
e^{(1-r/c)} & \text{if } c \le r
\end{cases}
\end{equation}
and $p_n$ is the modified precision for n-grams of size $n$,
\begin{equation}
p_n
=
\frac{
\sum_{\text{ngram}}
\min\bigl(
\mathrm{Count}_{\mathrm{cand}}(\text{ngram}),
\mathrm{Count}_{\mathrm{ref}}(\text{ngram})
\bigr)
}{
\sum_{\text{ngram}}
\mathrm{Count}_{\mathrm{cand}}(\text{ngram})
}.
\end{equation}

chrF++ measures character n-gram precision and recall, CodeBLEU augments n-gram overlap with syntactic and semantic program signals, and ROUGE-L measures longest-common-subsequence overlap rather than n-gram overlap. We chose this family for its ability to identify direct matches between generated and ground-truth code; the metrics are cheap and give a rough indication of how close a model's output is to the reference. They are limited, however, in that they cannot capture the geometric relations between elements: a high score may reflect only shared boilerplate, while a correct diagram built from different Asymptote functions may be penalized. Since Asymptote code is highly formulaic and reuses many of the same functions, we retain these metrics as coarse indicators of consistency rather than measures of geometric correctness.

\paragraph{Image-Based Metrics}
For visual fidelity, we use Learned Perceptual Image Patch Similarity (LPIPS) ~\cite{zhang2018unreasonable}, which compares deep feature representations of the rendered diagrams. It is computed as:

\begin{equation}
d(x, x_0)
=
\sum_l \frac{1}{H_l W_l}
\sum_{h,w}
\left\|
w_l \odot
\left(
\hat{\phi}_{l,hw}^{(x)}
-
\hat{\phi}_{l,hw}^{(x_0)}
\right)
\right\|_2^2
\end{equation}

where $\hat{\phi}_{l,hw}^{(x)}$ is the normalized activation tensor
from layer $l$ of image $x$ at spatial position $(h, w)$, $H_l$ and
$W_l$ are the height and width of the feature map at layer $l$, and
$w_l$ is a learned weight. Lower values indicate greater similarity.

\paragraph{VLM-Based Metrics}
We use GPT-5.4-mini as a vision-language judge, scoring each generated
diagram against the ground truth on a scale from 0 to 10, which we rescale to 0-100 for consistency with our other metrics. The full
prompt is given in Appendix~\ref{app:prompts}.

\paragraph{Compilation Metrics}
We report a Compile Success Rate, the fraction of generated Asymptote programs that render to an image without error. It captures only whether a model's output is usable, not whether the resulting diagram
is geometrically correct, and so serves as a lower bound on diagram quality.

\section{Dataset}
\subsection{Curation}
\begin{figure*}[htbp]
  \centering
  \includegraphics[width=\textwidth]{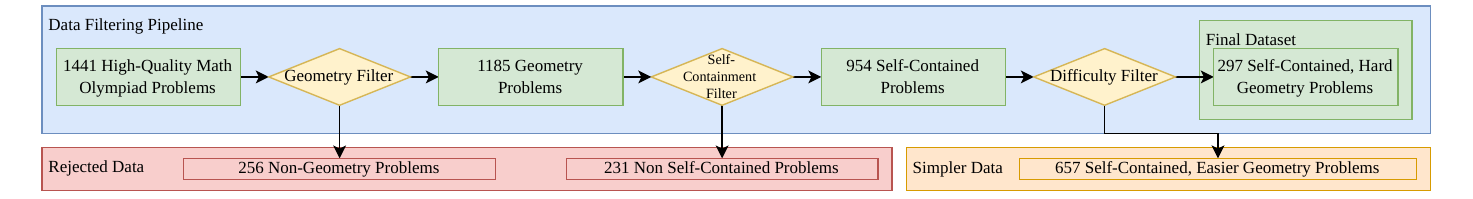}
  \caption{Overview of the Problems Filtered during the Dataset Curation Process. We web-scraped a total of 1441 raw datapoints, before filtering out the problems that were not geometric and not self-contained.}
  \label{fig:diagram4}
\end{figure*}

Figure \ref{fig:diagram4} provides an overview of our filtering pipeline during the dataset curation process. 

\subsection{Breakdown}
A full breakdown of our dataset according to the original problem source and difficulty score is as follows:
\label{app:dataset-breakdown}
\begin{table*}[h!]
    \centering
    \caption{Final Dataset Breakdown based on Difficulty}
    \label{tab:baselineresults}
    \begin{tabular}{lcccccccccccc}
        \toprule
        \textbf{Dataset} & \textbf{1} & \textbf{2} & \textbf{3} & \textbf{4} & \textbf{5} & \textbf{6} & \textbf{7} & \textbf{8} & \textbf{9} & \textbf{10} & \textbf{Total} \\
        \midrule
        AMC & 0 & 14 & 29 & 91 & 95 & 84 & 51 & 15 & 0 & 0 & 379 \\
        AIME & 0 & 0 & 13 & 63 & 83 & 116 & 105 & 52 & 8 & 0 & 440 \\
        USAJMO & 0 & 0 & 0 & 2 & 5 & 6 & 3 & 5 & 0 & 0 & 21 \\
        USAMO & 0 & 0 & 1 & 3 & 3 & 5 & 14 & 15 & 6 & 1 & 48 \\
        IMO & 0 & 0 & 0 & 0 & 1 & 1 & 3 & 1 & 2 & 0 & 8 \\
        Others & 0 & 0 & 5 & 12 & 12 & 16 & 9 & 4 & 0 & 0 & 58 \\
        Total & 0 & 14 & 48 & 171 & 199 & 225 & 188 & 92 & 16 & 1 & 954 \\
        \bottomrule
    \end{tabular}
\end{table*}

\begin{figure*}[htbp]
  \centering
  \includegraphics[width=\textwidth]{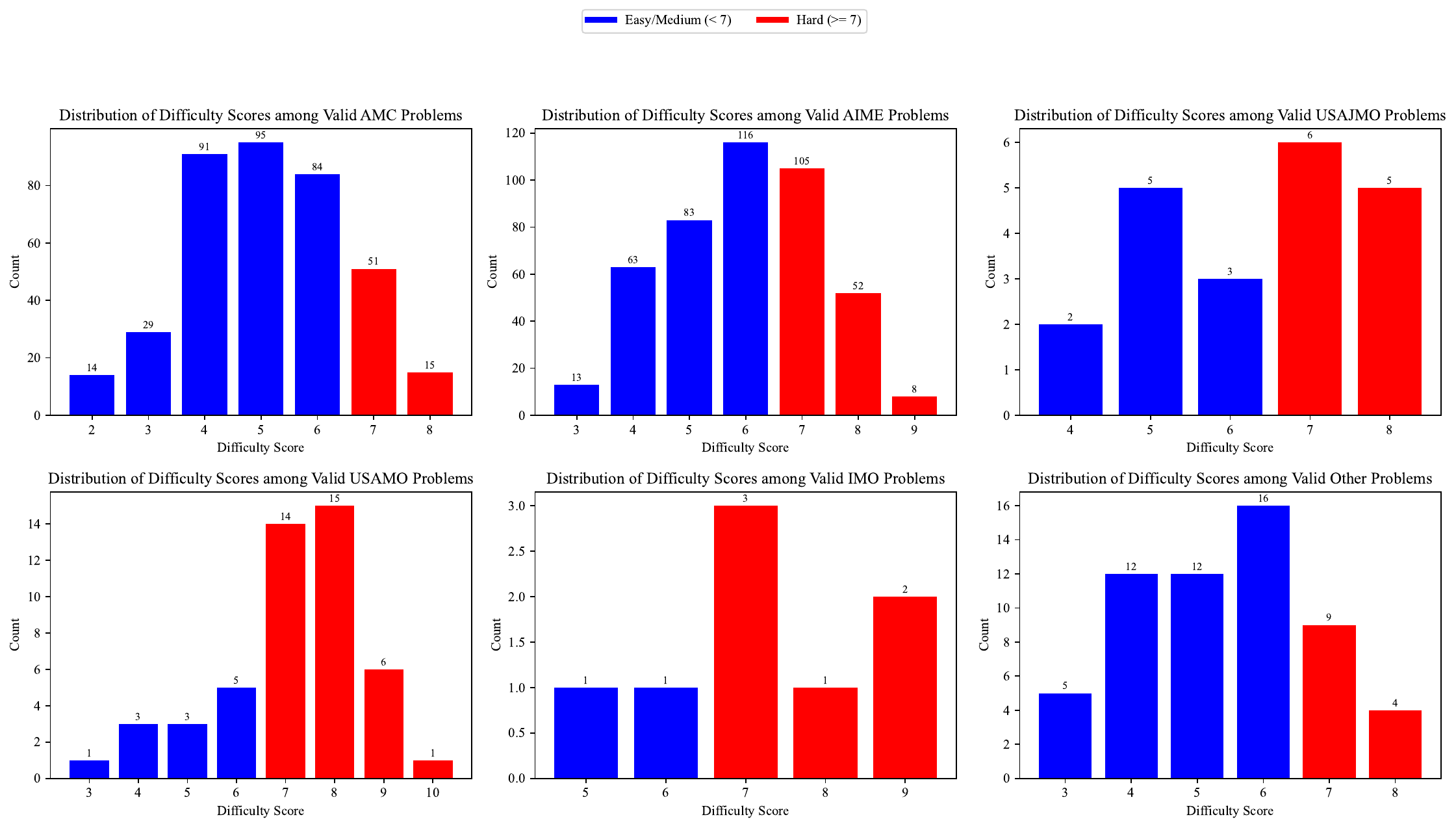}
  \caption{Breakdown of Problems by Difficulty Score}
  \label{fig:diagram5}
\end{figure*}

\begin{figure*}[htbp]
  \centering
  \includegraphics[width=\textwidth]{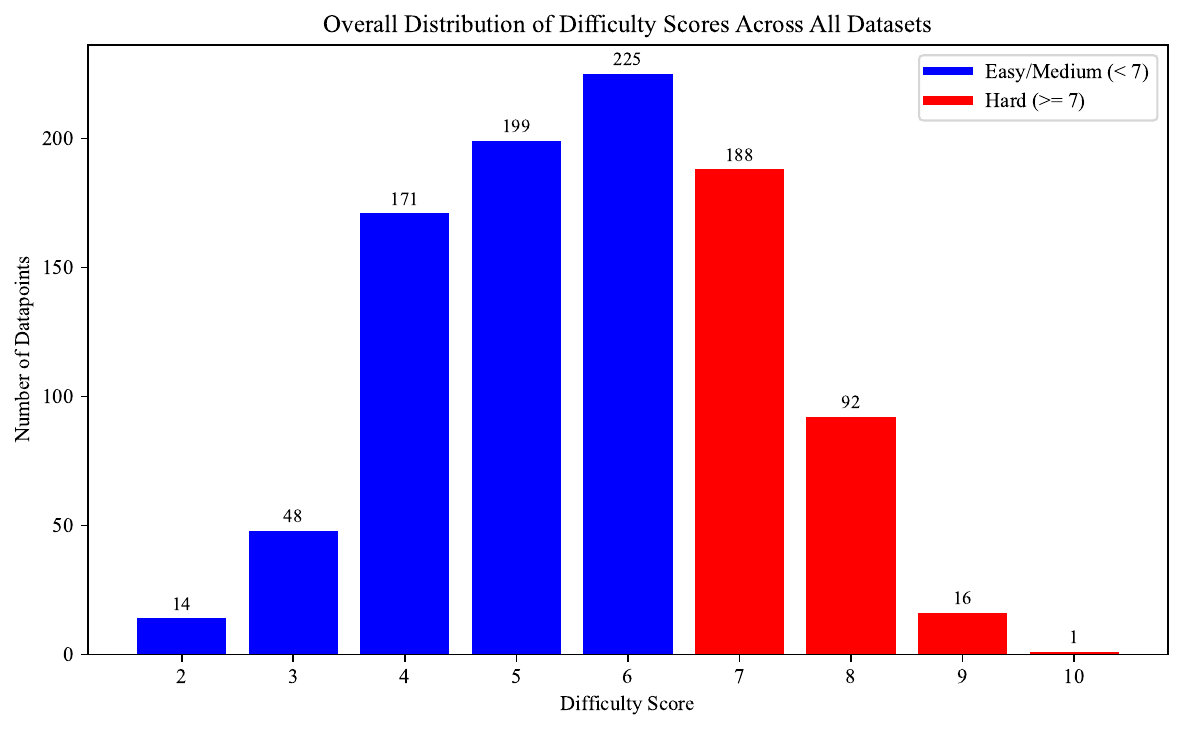}
  \caption{Breakdown of All Problems by Difficulty Score}
  \label{fig:diagram6}
\end{figure*}

\end{document}